\documentclass[sigconf,screen, natbib=true]{acmart}
\usepackage{graphicx}
\usepackage{subcaption}
\usepackage{algorithm}
\usepackage{algpseudocode}
\usepackage{enumitem}
\usepackage{multirow}
\usepackage{tabularx}
\usepackage{threeparttable}
\usepackage{color, xspace}
\usepackage{appendix}
\usepackage{balance}

\AtBeginDocument{%
	\providecommand\BibTeX{{%
			\normalfont B\kern-0.5em{\scshape i\kern-0.25em b}\kern-0.8em\TeX}}}
\copyrightyear{2025} 
\acmYear{2025}
\setcopyright{acmlicensed}\acmConference[KDD '25]{Proceedings of the 31st ACM
SIGKDD Conference on Knowledge Discovery and Data Mining V.2}{August 3--7,
2025}{Toronto, ON, Canada}
\acmBooktitle{Proceedings of the 31st ACM SIGKDD Conference on Knowledge
Discovery and Data Mining V.2 (KDD '25), August 3--7, 2025, Toronto, ON, Canada}
\acmDOI{10.1145/3711896.3737257}
\acmISBN{979-8-4007-1454-2/2025/08}

\begin{document}
\title[Put Teacher in Student's Shoes: Cross-Distillation for Ultra-compact Model Compression Framework]{Put Teacher in Student's Shoes: Cross-Distillation for Ultra-compact Model Compression Framework}
\author{Maolin Wang}
\authornote{Both authors contributed equally to this research.}
\authornote{All work performed during internship at AntGroup}
\affiliation{%
  \institution{City University of Hong Kong and AntGroup}
   \city{Hong Kong SAR and Hangzhou}
   \country{China}
  }
\email{morin.wang@my.cityu.edu.hk}

\author{Jun Chu}
\authornotemark[1]
\affiliation{%
  \institution{AntGroup}
  \city{Hangzhou}
  \country{China}
  }
\email{chujun.chu@antgroup.com}


\author{Sicong Xie}
\affiliation{%
  \institution{AntGroup}
  \city{Hangzhou}
  \country{China}
  }
\email{sicong.xsc@antgroup.com}

\author{Xiaoling Zang}
\affiliation{%
  \institution{AntGroup}
  \city{Hangzhou}
  \country{China}
  }
\email{zangxiaoling.zxl@antgroup.com}

\author{Yao Zhao}
\affiliation{%
  \institution{AntGroup}
  \city{Hangzhou}
  \country{China}
  }
\email{nanxiao.zy@antgroup.com}

\author{Wenliang Zhong}
\authornote{Corresponding Authors}
\affiliation{%
  \institution{AntGroup}
  \city{Hangzhou}
  \country{China}
  }
\email{yice.zwl@antgroup.com}

\author{Xiangyu Zhao}
\authornotemark[3]
\affiliation{%
  \institution{City University of Hong Kong}
  \city{Hong Kong SAR}
  \country{China}
  }
\email{xianzhao@cityu.edu.hk}

\renewcommand{\shortauthors}{Maolin Wang, et al.}
\begin{abstract}

In the era of mobile computing, deploying efficient Natural Language Processing (NLP) models in resource-restricted edge settings presents significant challenges, particularly in environments requiring strict privacy compliance, real-time responsiveness, and diverse multi-tasking capabilities. These challenges create a fundamental need for ultra-compact models that maintain strong performance across various NLP tasks while adhering to stringent memory constraints.
To this end, we introduce \textbf{E}dge ultra-l\textbf{I}te \textbf{BERT} framework (\textbf{EI-BERT}) with a novel cross-distillation method. EI-BERT efficiently compresses models through a comprehensive pipeline including hard token pruning, cross-distillation, and parameter quantization. Specifically, the cross-distillation method uniquely positions the teacher model to understand the student model's perspective, ensuring efficient knowledge transfer through parameter integration and the mutual interplay between models.
Through extensive experiments, we achieve a remarkably compact BERT-based model of only \textbf{1.91 MB} - the smallest to date for Natural Language Understanding (NLU) tasks. This ultra-compact model has been successfully deployed across multiple scenarios within the Alipay ecosystem, demonstrating significant improvements in real-world applications. For example, it has been integrated into Alipay's live Edge Recommendation system since January 2024, currently serving the app's recommendation traffic across \textbf{8.4 million daily active devices}.

\end{abstract}




\maketitle
\section{Introduction}
In recent years, Natural Language Processing (NLP) has made significant strides, particularly in developing compact and efficient models for resource-constrained environments such as mobile and edge computing devices~\cite{khurana2023natural,min2023recent,guo2019deep,vaswani2017attention,radford2018improving, wang2023large}. This progress has been driven by the increasing demand for AI solutions that are both powerful and efficient, capable of running on devices with limited computational resources. The challenge lies in balancing computational efficiency with model performance, a critical consideration for applications that require real-time responsiveness and strict privacy compliance~\cite{xu2023survey,menghani2023efficient,wang2023wireless,wang2023tensor,zhaok2021autoemb,zhao2021autodim}. 

Despite advancements in model compression techniques, current solutions still face significant challenges in fully addressing the complexities of real-world applications. These complexities include ensuring privacy through on-device operations, achieving real-time responsiveness, and supporting diverse multi-tasking functionalities~\cite{zhou2023opportunities,choudhary2020comprehensive,baccour2022pervasive,shi2020communication}. In real-world AI model deployment, advancements in model compression have significantly improved efficiency and personalization. Techniques such as Taobao's dynamic recommendation~\cite{song2022autoassign,zhaok2021autoemb}, IBM's transformer model compression~\cite{wang2023large}, and Amazon's edge device optimization demonstrate reduced model size with preserved accuracy and enhanced computational efficiency in limited-resource settings~\cite{wang2024large,gong2020edgerec, wang2022deep, kharazmi2023distill,lin2023autodenoise}. Nevertheless, despite their significance, these advancements do not fully address the broad spectrum of challenges associated with deploying AI language models in real-world scenarios~(like Alipay ecosystem). First, in environments like Alipay which are sensitive to privacy and handle financial data, there is a critical need for local AI processing to preserve user privacy, emphasizing the importance of developing models capable of operating on users' devices. Second, the need for real-time responsiveness in applications highlights the essential role of model efficiency, necessitating the precision engineering of models to achieve a response time of approximately 300 ms on local devices. Third, the necessity to support diverse multi-tasking functionalities~\cite{wang2022autofield,gao2023autotransfer} underscores the need for deploying specialized AI models~\cite{chen2022automated}, where an effective model delivery strategy is essential, prioritizing ultra-compact model sizes to enhance speed.

These requirements create a fundamental need: achieving ultra-compact model sizes while maintaining strong knowledge capabilities for diverse NLP tasks. To meet these constraints and needs in Alipay's mobile environment, an ultra-compact Natural Language Understanding (NLU) model \textbf{strictly limited to 4MB} that ensures both instant processing capabilities and robust performance must be developed.
However, current BERT model compression techniques, achieving only 15-20MB reductions~\cite{gong2020edgerec, mirzadeh2020improved, wang2022deep, kharazmi2023distill,gupta2022compression, xu2023survey} and lacking focus on the delivery of multiple online compressed models, present significant challenges for deployment in memory-constrained and multi-scenario environments like Alipay's. This disparity underscores the urgent need for an extreme compression framework to fulfill the stringent memory budgets of diverse mobile devices, thereby ensuring a personalized user experience.

Recognizing this unmet need, we introduce our \textbf{E}dge ultra-l\textbf{I}te \textbf{BERT} framework (\textbf{EI-BERT}), with a novel \textbf{cross-distillation method}. The EI-BERT framework efficiently compresses models through a series of steps, progressing through hard token pruning, cross-distillation, and parameter quantization. A pivotal aspect of this framework is our cross-distillation approach. This approach effectively puts the teacher model in the student model's shoes and ensures efficient and precise knowledge transfer through parameter integration and mutual interplay between the teacher and student models~\cite{chen2022automated}. It significantly enhances knowledge transfer, thereby improving model efficiency and robustness. Nevertheless, to further optimize the model's performance and deployment efficiency, EI-BERT culminates in post-training quantization, utilizing a module-wise approach~\cite{zhaok2021autoemb,zhao2021autodim} to preserve model integrity. 

We demonstrated the effectiveness of our EI-BERT framework through extensive experiments, achieving a remarkably compact BERT-based language model of only 1.91MB - the smallest to the best of our knowledge. This ultra-compact size enables seamless deployment on mobile devices while maintaining robust performance, making it a particularly valuable Alipay that demands both efficiency and versatility. Our work presents four key contributions:
\begin{itemize}[leftmargin=*]
\item We introduce a groundbreaking compression framework that reduces the model's memory footprint to approximately 1.91 MB in NLU scenarios, achieving a 99.5$\%$ reduction in size compared to BERT-base models. This ensures the model can be run and compiled on mobile devices while preserving essential performance metrics for real-time applications.
\item We propose innovative cross-distillation techniques that effectively transfer knowledge from larger, complex models to ultra-light, streamlined models in Natural Language Understanding (NLU) scenarios.
\item We rigorously validate our methodology across diverse datasets, including public CLUE benchmark~\cite{xu2020clue} and proprietary data from Ant Group, demonstrating the robustness, adaptability, and high performance of our approach. 
\item We successfully deployed our framework in multiple real-world NLU scenarios within Alipay, an app with over a billion users, achieving excellent online results. To the best of our knowledge, our 1.91MB NLU model is the most compact BERT-based language model deployed in a real-world industrial setting.
\end{itemize}

\textbf{Significance.} Our work sets a new milestone in model compression and efficiency, paving the way for further research and development aimed at enhancing AI accessibility and effectiveness across a diverse range of edge-based applications. This significant advancement not only optimizes performance but also reduces resource consumption, thereby facilitating the broader deployment of AI technologies and increasing their real-world impact in various mobile environments.

\section{Methodology}

\begin{figure*}[htbp]
\centering
\includegraphics[width=0.9\textwidth]{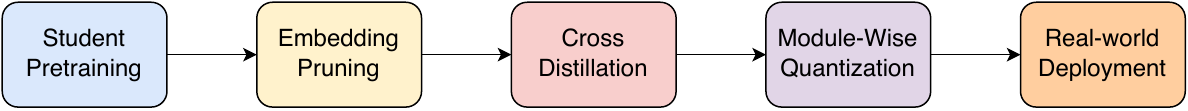}

\caption{
The proposed ultra-compact model compression framework.
}
\label{fig:pipeline}
\end{figure*}
\begin{figure*}[htbp]
\centering
\includegraphics[width=\textwidth]{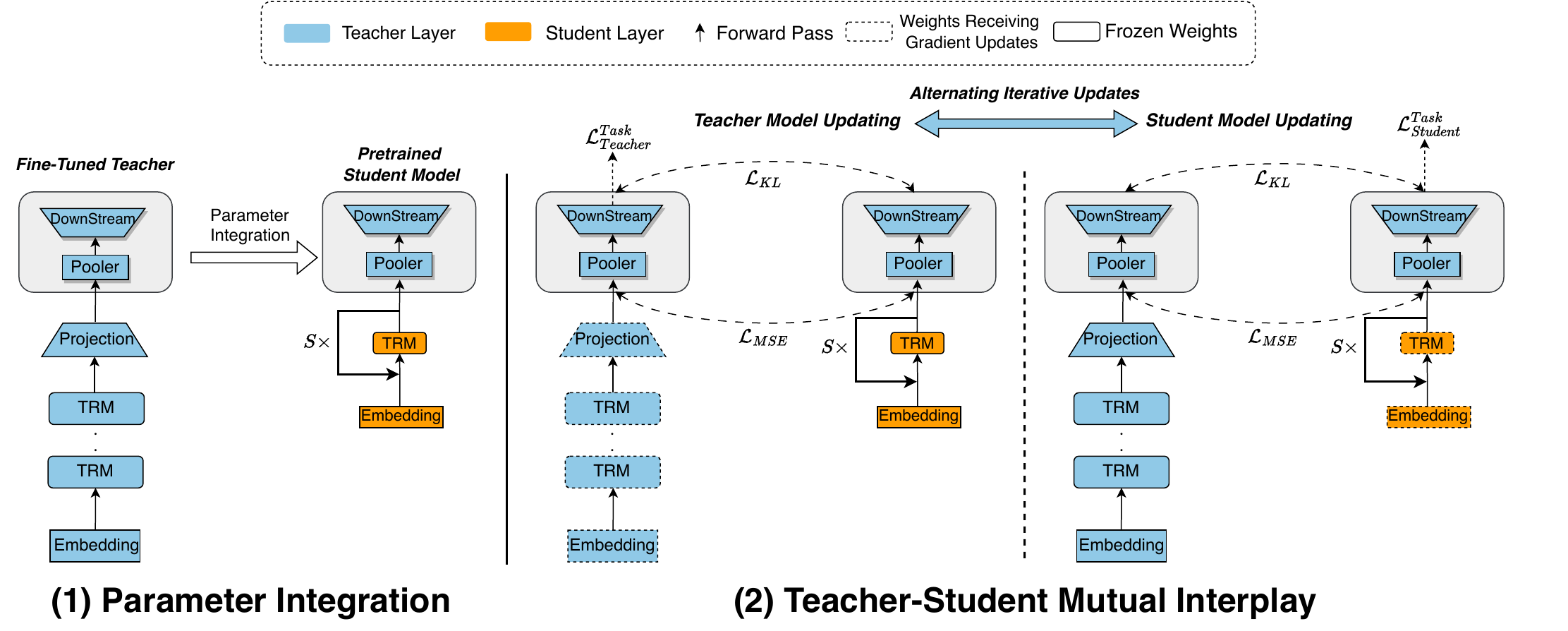}
\vspace{-10pt}
\caption{Cross-distillation featuring parameter integration and teacher-student mutual interplay advances beyond conventional methods by integrating teacher components into an extremely compact student model, ensuring compatibility and effective knowledge transfer at every training step.
}
\vspace{-10pt}
\label{fig:crossdistillation}
\end{figure*}
This subsection details the framework for transferring knowledge from teacher models to ultra-compact student models. As shown in Figure~\ref{fig:pipeline} and Figure~\ref{fig:crossdistillation}, EI-BERT begins with student pre-training to simplify architectures and eliminate redundancies. Cross-layer distillation then transfers critical knowledge, followed by INT8 quantization—a mobile-friendly method widely supported on devices—to minimize storage. Designed for deployment in Alipay (serving over 1B users), this framework ensures compatibility while achieving extreme compactness.

\subsection{Student Pre-training}
To achieve the ultra compression of NLP models, the selection of student models is of paramount importance. We use tiny ALBERT-2~\cite{lan2019albert} as our basic pretrained student model architecture containing factorized embedding parameterization and cross-layer parameter sharing. Additionally, we discuss in detail how to initialize the student models in Section~\ref{sec:experiments}.

\subsection{ Hard Token Pruning}

The high parameter allocation to token embeddings, as seen with 44.7\% in ALBERT-tiny$_2$~\cite{lan2019albert} and 36.6\% in TinyBERT$_2$~\cite{jiao2019tinybert}, underscores the storage burdens of lightweight transformer models. Existing methods~\cite{kim2021length,kim2022learned,wang2021spatten} mainly focus on inference computation cost and memory footprint, with limited certainty regarding model storage size. In response, we propose a hard token pruning strategy. 
We define the attention probability between tokens $\mathrm{x}_i$ and $\mathrm{x}_j$ from head $h$ as $\mathbf{A}^{(h)}(\mathrm{x}_i, \mathrm{x}_j)$:
\begin{equation}
\mathbf{A}^{(h)}\left(\mathrm{x}_i, \mathrm{x}_j\right) = \operatorname{softmax}\left(\frac{\mathbf{x}^T \mathbf{W}_q^T \mathbf{W}k \mathbf{x}}{\sqrt{d_k}}\right){(i, j)},
\end{equation}
where $\mathbf{x}$, $\mathbf{W}_q$, $\mathbf{W}_k$, and $d_k$ denote the input token embeddings, the query weight matrices and key weight matrices, and the dimensionality of the key vectors, respectively.

For token pruning, it is necessary to establish a metric for identifying unimportant tokens. Consequently, attention probabilities are computed for each token within a corpus of sentences to determine their respective importance. Inspired by~\cite{goyal2020power,kim2021length,kim2022learned}, we calculate the token importance $I(w)$ using the following approach:
\begin{equation}
I(w) = \sum_{k=1}^{K} \frac{1}{n_k} \sum_{h=1}^{H} \sum_{j=1}^{n_k} \mathbf{A}^{(h)}(w, x_j),
\end{equation} 
with $K$, $H$, and $n_k$ representing the number of layers, sentence count, attention heads, and the length of the $k$-th sentence, respectively. Leveraging a top-$k$ selection based on importance scores for token pruning, our strategy significantly enhances computational efficiency and reduces storage requirements, thereby addressing the critical challenges of model size and complexity. Practically, we use the attention mechanism of the last layer of students for importance scoring calculation.


\subsection{Cross Distillation}

When language models are distilled to an ultra-compact size, they encounter two primary challenges:

\begin{itemize}[leftmargin=*]
    \item \textbf{Capacity Disparity Between Teacher and Ultra-compact Student:} The capacity gap~\cite{zhou2021bert,yang2022sparse,zhang2022minidisc} in ultra-compact compression scenarios between teacher and student models poses a significant hurdle. The gap, stemming from the variance in information processing capabilities, can severely hinder distillation.
    
    \item \textbf{Adaptation Gaps in Knowledge Transfer:} Traditional distillation methods overlook the unique learning needs and environmental contexts of ultra-compact models. They employ static teacher models~\cite{mirzadeh2020improved,zhou2021bert,liang2023less} providing fixed instructional content that doesn’t adapt to student progress or specific challenges. This one-size-fits-all approach risks suboptimal outcomes, with students merely replicating teacher outputs without internalizing or adapting knowledge for operational needs
\end{itemize}

Due to these two challenges, traditional distillation methods are difficult to apply in ultra-compact scenarios. To address the difficulties that extremely compact student models face when assimilating complex knowledge from large teacher models, we introduce a novel distillation method called cross-distillation. This approach ensures efficient and precise knowledge transfer by effectively placing the teacher in the student’s position. 
The core principle behind cross-distillation is to enhance the knowledge absorption and utilization efficiency of ultra-compact student models through integration and dynamic adaptation. Cross-distillation enables the teacher to find an optimal matching space from the student’s perspective by allowing the advanced teacher model to adapt dynamically to the student’s capabilities. This dynamic interplay enhances knowledge transfer, allowing the student model to maximally benefit from the teacher’s expertise.
As shown in Figure~\ref{fig:crossdistillation}, cross-distillation comprises two primary steps: Parameter Integration step and Teacher-Student Mutual Interplay step.

\subsubsection{\textbf{Parameter Integration}}
Following the fine-tuning of the teacher model, we directly utilize the teacher’s downstream task-specific non-transformer layers for the student model rather than training new ones. The motivation for directly integrating these task-specific parameters is to reduce the learning difficulty for the student model, enabling it to quickly acquire task-related knowledge and significantly enhance its learning efficiency. Considering the architecture of language models~\cite{kim2021length, wang2019language, petroni2019language}, the discriminative information contained in the last layer of the teacher model encompasses substantial task-specific knowledge, as highlighted by~\cite{jawahar2019does}. Inspired by~\cite{chen2022knowledge}, we assume that fine-tuning the teacher model sufficiently trains these downstream layers, making them ready for integration into the student model. This integration significantly enhances the learning efficiency and pathway for the student model.

\subsubsection{\textbf{Teacher-Student Mutual Interplay}}
The motivation for the teacher-student mutual interplay is to ensure the efficiency and accuracy of knowledge transfer via mutual dynamic adaptation. Combining the final layer outputs and intermediate hidden feature representations of both teacher and student models and applying two different loss functions to refine the alignment process enhances the models' robustness and efficiency.
We achieve this by combining the outputs of the final downstream layer and the intermediate hidden feature representations from both the student and teacher models. For aligning the downstream logits outputs, Kullback-Leibler (KL) divergence is applied, and mean squared error (MSE) loss is used for the alignment of intermediate hidden feature representations~\cite{kim2021comparing}. These loss functions are strategically selected to enhance the knowledge transfer mechanism, ensuring an efficient and precise transfer of critical insights. This step is divided into cross-updates of the teacher and the student.

\noindent \textbf{Teacher Model Updating.}
First, the teacher model is updated by leveraging both the student model and the specific task at hand. The motivation here is that by continuously updating the teacher model during training, we can enhance its guiding ability, making it more effective in interacting with and fostering the growth of the student model.
Through teacher updating, our objective is to enhance the teacher model’s capacity for proficient guidance and interaction with the student models.
Specifically, we utilize a relatively small learning rate, $\lambda_{1}$. The specific loss function is defined as:
\begin{equation}
\mathcal{L}_{\text{teacher}} = \mathcal{L}_{\text{task}}^{\text{teacher}} + \beta{1} \cdot \mathcal{L}_{\text{mse}}(f^{(h)}_{t}, f^{(h)}_{s})+\beta_{2} \cdot \mathcal{L}_{\text{KL}}(f_{t}, f_{s})
\end{equation}
Here, $\mathcal{L}_{\text{teacher}}$ represents the comprehensive loss for the teacher model, comprising the task-specific loss, $\mathcal{L}_{\text{task}}^{\text{teacher}}$, and the MSE loss $\mathcal{L}_{\text{mse}}$ calculated between the hidden feature representations of the teacher and student models before the final pooling layer and classification head. These hidden feature representations are denoted as $f^{(h)}_{t}$ and $f^{(h)}_{s}$ for the teacher and student models, respectively. And $f_{t}$ and $f_{s}$ refer to the downstream output logit of the teacher and student models, respectively. 

\noindent \textbf{Student Model Updating.}
In contrast, the student model then utilizes a larger learning rate to be updated, $\lambda_{2}$.
This ensures that the student model not only learns the final outputs but also absorbs the complex internal representations of the teacher model, thereby achieving comprehensive knowledge transfer.
The loss function for the student model combines the task-specific loss with dual instances of the MSE loss, which can be described as:
\begin{equation}
\mathcal{L}_{\text{student}} = \mathcal{L}_{\text{task}}^{\text{student}} + \beta_{1} \cdot \mathcal{L}_{\text{mse}}(f^{(h)}_{t}, f^{(h)}_{s}) + \beta_{2} \cdot \mathcal{L}_{\text{KL}}(f_{t}, f_{s})
\end{equation}
These loss structures are fundamental to our cross-distillation process, ensuring that the student model learns not only final outputs but also intricate internal representations of the teacher.

Following the approach outlined previously, we iteratively update the student model and teacher model. {This iterative process entails a dynamic exchange of information between the teacher and student models, resulting in continuous updates driven by their respective losses.} As the training progresses, the student model steadily refines its understanding and hones its representations, progressively converging towards acquiring the nuanced knowledge the teacher imparted. 

Moreover, when tackling specific tasks, the loss function utilized in cross-distillation is tailored to fit the requirements of those tasks. The Mean Squared Error loss, denoted as $L_{MSE}$, remains constant and unaltered throughout the distillation process. Similarly, the Kullback-Leibler divergence loss, represented as $L_{KL}$, does not change significantly across various classifications, including generative tasks. This consistency is mainly because these tasks typically culminate in a softmax layer, which standardizes the output distribution. Finally, the task-specific loss indicated as $L_{task}$ is modified to align with the original loss functions corresponding to each task. 

\subsection{Module-wise Quantization}

Building on initial vocabulary pruning and model distillation, we further enhance efficiency using post-training quantization~\cite{liu2021post}. Different from conventional matrix-wise methods~\cite{mishchenko2019low,wu2022noisytune,nahshan2021loss}, our quantization approach minimizes error holistically across interrelated layers within the model. 

Specifically, we aim to minimize discrepancies between quantized and full-precision outputs of modules. Considering our model is partitioned into $N$ modules following compression, the objective function for the $n$-th module in our model, which encompasses multiple transformer layers, is formulated in the following format:
\begin{equation}
    \min_{\mathbf{w}_n, \mathbf{s}_n} \ell^{(n)}=\sum_{l \in\left[l_n, l{n+1}\right)}\left|\hat{\boldsymbol{f}}_l-\boldsymbol{f}_l\right|^2
\end{equation}  
where $\ell^{(n)}$ is the reconstruction error for the $n$-th module, $\hat{\boldsymbol{f}}_l$ denotes the quantized output of the $l$-th layer within the module, $\boldsymbol{f}_l$ is the corresponding full-precision output, and $\mathbf{w}_n$ and $\mathbf{s}_n$ is the learnable quantization weights and step sizes for the module, respectively. Specifically, the choice of using 8-bit quantization has been made, and the selection of the step size $\mathbf{s}_n$, as a critical parameter substantially enhances the precision of our quantized values, optimizing model efficiency and accuracy.
To formalize our quantization process, we define the quantization function as follows:
\begin{equation}
\hat{\mathbf{w}_n}=\mathcal{Q}_b(\mathbf{w}_n) = \mathbf{s}_n \cdot \Pi_{\Omega(8)}\left(\frac{\mathbf{w}_n}{ \mathbf{s}_n}\right)
\end{equation}
where $\Omega(8)=\left\{-2^{7}+1, \ldots, 0, \ldots, 2^{7}-1\right\}$ is the set of $8$-bit integers, and $\Pi{(\cdot)}$ is the projection function that maps the input to its closest integer. We make particular efforts to ensure that the full dynamic range of the neural network’s parameters is precisely and comprehensively captured within the limitations imposed by the reduced lower-bit numerical representation, thereby minimizing potential quantization errors and preserving the accuracy, consistency, and overall integrity of the model’s output predictions.
Importantly, our decision not to use lower-bit precision or mixed-precision techniques is primarily motivated by the increasing availability and recent advancements in industrial-grade mobile hardware, which now provides robust support for efficient 8-bit quantization.

\section{Experiments}
\label{sec:experiments}
In this section, we systematically evaluate our proposed framework, with a particular focus on the cross-distillation technique.

\subsection{Pretrained Corpus}
~\label{appendix:pre1}
For our student model's pretraining, we chose a configuration that optimizes efficiency and effectiveness. 
We exclusively used the CLUECorpus2020-small dataset~\cite{xu2020clue}~(14GB of Chinese text) for general comprehensive language understanding and included the Alipay E-commerce Corpus~(5.8GB, 2.1 billion tokens) for our internal comprehensive language understanding. Various fundamental statistics for these datasets are presented in Table~\ref{tab:cluecorpus} and Table~\ref{tab:alipaycorpus}.

\subsection{Datasets}
Our framework evaluation employs two categories of datasets: public benchmarks and internal datasets. For public benchmarks, experiments are conducted on the \textbf{CLUE Benchmark}~\cite{xu2020clue}, which evaluates Chinese language understanding through Sentence Pair tasks and complex Machine Reading Comprehension (MRC) tasks. Details are shown in Table~\ref{tab:clue}. The internal Alipay Datasets, in contrast, encompass diverse tasks including Named Entity Recognition evaluated using F1-score (identifying 29 major entity types from Alipay's marketing and financial datasets), Classification tasks assessing 31 primary and 108 secondary categories (using accuracy as the primary metric), and Similarity tasks crucial for determining likeness between different data points. The detailed description can be found in Appendix Sec.~\ref{appendix:Alipay}.

\subsection{Implementations}
Our implementation is based on PyTorch\footnote{\url{https://pytorch.org}} and Huggingface\footnote{\url{https://github.com/huggingface/transformers}}.
The student model for language understanding employs 128-dim embeddings and 1024 intermediate layers. Hyperparameter searches covered batch sizes (8-32), student/teacher learning rates (1e-4-5e-5 / 1e-7-1e-6), and distillation parameters (beta1/beta2:1-10), trained on 8×V100 GPUs (<5 days total, 8-10 hours for cross-distillation). Quantization and token pruning maintain low overhead with multi-task model reuse. For generation tasks, GPT-2 (12 layers, 12 heads, hidden dim 768/1024) used batch size 32 and sequence length 512 on 8×A100 GPUs (80GB), with beta1/beta2 extended to 0.5-5.0. 

\begin{table}[!t]
\centering
\caption{Summary of CLUECorpus2020-small}
\label{tab:cluecorpus}
\resizebox{0.45\textwidth}{!}{%
\begin{tabular}{@{}lll@{}}
\toprule
\textbf{Genre} & \textbf{Description} & \textbf{Size/Words} \\
\midrule
News & Crawled from We Media platforms & 3 billion words \\
WebText & Questions and answers from forums & 410 million Q\&As \\
Wikipedia & Chinese content from Wikipedia & 0.4 billion words \\
Comments & Collected from E-commerce websites & 0.8 billion words \\
\midrule
\textbf{Total} & \textbf{Combined size of all sub-corpora} & \textbf{4.61 billion words} \\
\bottomrule
\end{tabular}%
}
\end{table}

\begin{table}[!t]
\centering
\caption{Summary of Alipay Data}
\label{tab:alipaycorpus}
\resizebox{0.45\textwidth}{!}{%
\begin{tabular}{@{}lll@{}}
\toprule
\textbf{Category} & \textbf{Description} & \textbf{Size} \\
\midrule
Financial Data & Alipay financial-related data & 0.3 billion words \\
Marketing Data & Alipay marketing data & 0.2 billion words \\
\midrule
\textbf{Total} & \textbf{Combined size } & \textbf{ 0.5 billion words} \\
\bottomrule
\end{tabular}%
}
\end{table}

\begin{table}[!t]
\centering
\caption{Summary of CLUE Benchmark Tasks}
\label{tab:clue_tasks}
\resizebox{0.466\textwidth}{!}{%
\begin{tabular}{@{}llllll@{}}
\toprule
Corpus & Train & Dev & Test & Metric & Source \\
\midrule
\multicolumn{6}{c}{\textbf{Single-Sentence Tasks}} \\
\midrule
TNEWS & 53.3k & 10k & 10k & acc. & news title and keywords \\
IFLYTEK & 12.1k & 2.6k & 2.6k & acc. & app descriptions \\
CLUEWSC2020 & 1,244 & 304 & 290 & acc. & Chinese fiction books \\
\midrule
\multicolumn{6}{c}{\textbf{Sentence Pair Tasks}} \\
\midrule
AFQMC & 34.3k & 4.3k & 3.9k & acc. & online customer service \\
CSL & 20k & 3k & 3k & acc. & academic (CNKI) \\
OCNLI & 50k & 3k & 3k & acc. & 5 genres \\
\midrule
\multicolumn{6}{c}{\textbf{Machine Reading Comprehension Tasks}} \\
\midrule
CMRC 2018 & 10k & 3.4k & 4.9k & Exact Match. & Wikipedia \\
ChID & 577k & 23k & 23k & acc. & novel, essay, and news \\
C$^3$ & 11.9k & 3.8k & 3.9k & acc. & mixed-genre \\
\bottomrule
\end{tabular}
}
\label{tab:clue}
\end{table}

\begin{table*}[t]
\centering
\caption{Overall Performance Comparison on Sentence and MRC Tasks}
\vspace{-5pt}
\label{tab:combined_performance}
\begin{tabular}{@{}l|ccccccc|ccccc@{}}
\toprule
&\multicolumn{7}{c|}{ \textbf{Sentence Tasks} } & \multicolumn{4}{c}{\textbf{MRC Tasks}} \\
\midrule
{\textbf{Models}} & \textbf{TNEWS} & \textbf{IFLY-TEK} & \textbf{WSC-2020} & \textbf{AFQMC} & \textbf{CSL} & \textbf{OCNLI} & \textbf{Avg} & \textbf{CMRC-2018} & \textbf{CHID} & \textbf{C3} & \textbf{Avg} \\
\midrule
BERT$_{\text{base}}$ (Teacher)  & 57.47 & 59.22 & 82.23 &73.98&80.93&75.49&71.53 &69.54 &78.35 &68.34&72.08 \\
\midrule
TinyBERT$_{4}$    &54.02&45.81&70.07&69.07&75.07&69.59&63.94  &44.82&50.15&46.89&47.29 \\
ALBERT$_{4}$      &53.35&48.71&63.38&69.92&74.56&65.12&62.51 &43.21&48.76&46.52&46.16 \\
TinyBERT$_{2}$    &51.08&47.78&64.70&69.23&70.10&66.85&61.62 &41.65&44.28&46.21&44.05 \\
ALBERT$_{2}$ (Student) &51.87&49.71&63.49&68.86&69.16&63.69&61.13 &40.51&42.13&46.04&42.89 \\
\midrule
TA (TinyBert4 TA) &50.97&48.00&59.50&64.50&67.50&61.50&60.33 &43.50&48.20&45.90&45.87 \\
TA (ALBERT4 TA)  &52.32&49.50&60.70&65.30&68.20&62.30&61.13 &43.80&48.50&46.10&46.13 \\
Meta-KD          &52.71&50.00&61.20&65.70&68.70&62.80&61.95 &44.10&48.80&46.30&46.40 \\
\midrule
KD$_{\text{stu}}$ &51.02&53.90&57.24&69.34&69.07&64.37&60.82 &42.87&47.14&45.62&45.21 \\
PI-KD$_{\text{stu}}$ &52.08&54.52&63.48&69.11&71.80&66.20&62.87 &44.29&51.22&46.17&47.23 \\
\textbf{CrossKD$_{\text{stu}}$} &54.92&57.14&66.12&70.16&72.50&67.53&64.73 &48.49&55.79&47.67&50.65 \\
\textbf{CrossKD-TP$_{\text{stu}}$}&54.65&56.67&66.03&70.20&72.32&67.02&64.48 &48.31&55.35&47.48&50.38 \\
\textbf{EI-BERT}          &53.98&55.98&65.58&69.65&71.89&66.71&63.97 &47.76&55.23&47.12&50.04 \\ 
\bottomrule
\end{tabular}%
\vspace{-5pt}
\end{table*}
\begin{table}[!t]
\caption{Experiment Result On Alipay Dataset}
\vspace{-5pt}
\label{tab:alipaydata}
\begin{tabular}{c|ccc|c}
\hline
Model                     & NER   & Classification & Similarity                                      & AVG   \\ \hline
BERT$_{base}$(Teacher) & 84.82 & 75.13      &  94.23 & 84.73 \\
ALBERT$_{2}$(Student)             & 77.23 & 71.78      & 91.65                                     & 80.22 \\\hline\hline
KD$_{stu}$                      & 77.78 & 71.03      & 91.03                                     & 79.95 \\
PI-KD$_{stu}$     & 79.67 & 72.28      & 92.10                                      & 81.35 \\
\textbf{CrossKD$_{stu}$ }                     & 83.45 & 73.98      & 92.78                                     & 83.40 \\\hline\hline
\textbf{CrossKD-TP$_{stu}$}                     & 83.20 & 73.60      & 92.58                                  & 83.13 \\
\textbf{EI-BERT}         & 82.97 & 73.00      & 92.30                                     & 82.76 \\\hline
\end{tabular}
\vspace{-15pt}
\end{table}

\subsection{Baselines}
The baseline models are structured as follows: For Natural Language Understanding (NLU), we adopt a finetuned BERT${base}$~\cite{devlin2018bert} as the teacher model. Student models align with ALBERT$2$~\cite{lan2019albert}, including variants optimized through vanilla knowledge distillation (KD${stu}$), parameter integration (PI-KD${stu}$), and cross-distillation (CrossKD$_{stu}$). Our proposed EI-BERT extends CrossKD with token pruning and 8-bit quantization. Comparisons include Chinese-optimized TinyBERT~\cite{jiao2019tinybert} and ALBERT variants (ALBERT$_4$/ALBERT$_2$).
Additionally, we benchmark against task-aware distillation (TED~\cite{liang2023less}), multi-step Teacher-Assistant methods (TA~\cite{mirzadeh2020improved} with TinyBERT$_4$ and ALBERT$_4$), and adaptive Meta KD~\cite{zhou2021bert}. A more detailed description of the baseline is provided in Appendix Sec~\ref{sec:base}.

\subsection{Overall Performance on NLU Tasks}

\begin{figure}[!t]
    \centering
    \begin{subfigure}[b]{0.23\textwidth}
        \centering
        \includegraphics[width=\textwidth]{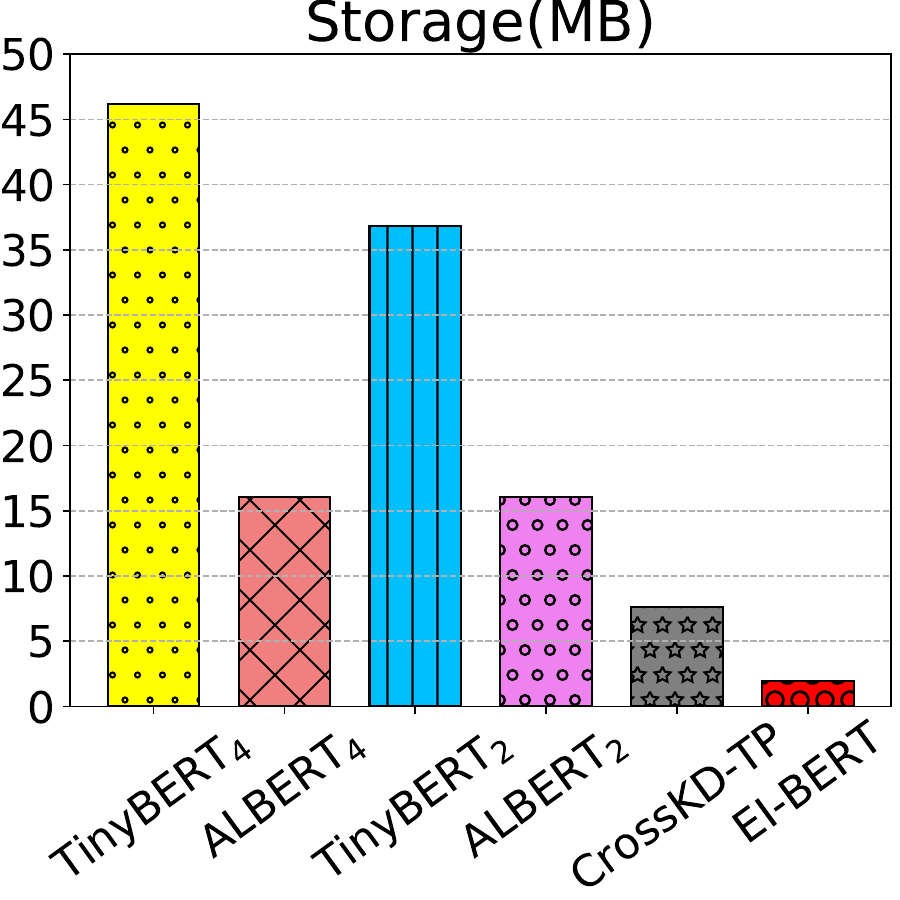} 
        \vspace{-3pt}
        \caption{Model storage.}
        \label{fig:storage}
    \end{subfigure}
    \hfill 
    \begin{subfigure}[b]{0.23\textwidth}
        \centering
        \includegraphics[width=\textwidth]{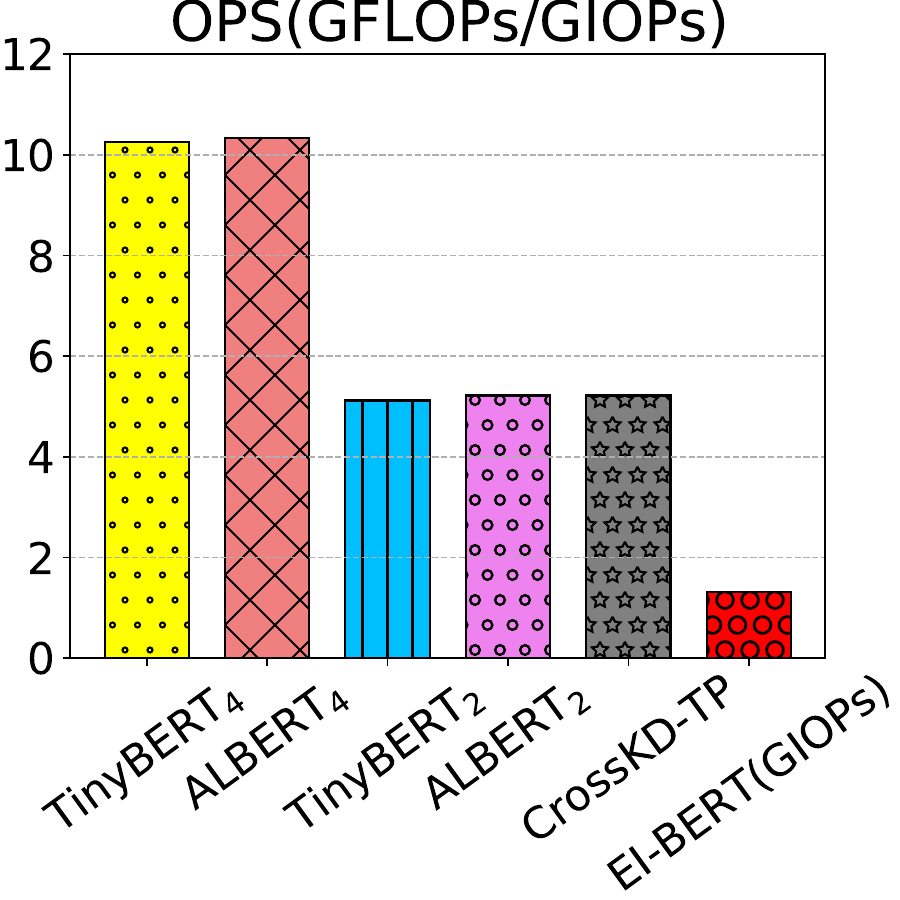} 
        \vspace{-3pt}
        \caption{OPs numbers. }
        \label{fig:flopsin}
    \end{subfigure}
    \caption{Overall Model Efficiency from Storage, and Model OPerations~(OPs). It is worth noting that, in~\ref{fig:flopsin}, EI-BERT is evaluated in terms of faster Integer Operations~(IOPs), whereas other models are assessed using Floating Point Operations~(FLOPs).}
    \label{fig:flops}
\vspace{-15pt}
\end{figure}

As demonstrated in Table~\ref{tab:combined_performance} and Figure~\ref{fig:flops}, the EI-BERT, with only 1.92MB in storage and 1.31 GIOPS, outperforms the TinyBERT4 model in the CLUE benchmark's sentence pair tasks, achieving an average score of 63.97 compared to TinyBERT$_{4}$'s 63.94, TinyBERT$_{2}$'s 61.62 and ALBERT$_{2}$ (Student)'s 61.13. 
The superior performance of CrossKD can be attributed to its effective knowledge transfer mechanism, which leverages insights from both teacher and student models to enhance learning outcomes. Unlike TinyBERT4, TinyBERT$_{2}$ and ALBERT$_{2}$~(Student), which do not incorporate such a comprehensive knowledge distillation approach, CrossKD's methodology significantly improves the model's ability to understand and process sentence pairs. 
This evidence advocates for adopting CrossKD in scenarios where maximizing performance without compromising computational resources is paramount.

Table~\ref{tab:combined_performance} also demonstrates that the EI-BERT model achieves an average score of 50.04 on more challenging Machine Reading Comprehension (MRC) tasks. This performance is marginally lower than that of CrossKD$_{\text{stu}}$, which scores 50.65, yet it is noteworthy given EI-BERT's significant reduction in model parameters. 
The model's ability to retain high performance, although with fewer parameters, is attributed to its comprehensive optimization strategy that incorporates cross-distillation, quantization, and pruning. This outcome confirms that models can maintain high-performance levels on complex tasks. EI-BERT's success underscores the importance of a multifaceted optimization approach in addressing demanding tasks like MRC. It demonstrates the feasibility of preserving model effectiveness amidst resource constraints, particularly through storage reduction, as elaborated in Figure~\ref{fig:flops}.

 Moreover, as shown in Table~\ref{tab:combined_performance}, our CrossKD significantly outperforms both Meta KD and TA methods. Specifically, on sentence-level tasks, CrossKD achieves an average score of 64.73, surpassing Meta KD (61.95) and TA methods (60.33 for TinyBert4, 61.13 for ALBERT4) by a notable margin. The performance gap is even more pronounced in MRC tasks, where CrossKD reaches 50.65 average accuracy, showing substantial improvements over Meta KD (46.40) and TA variants (45.87 and 46.13). This superior performance stems from our direct knowledge transfer strategy that avoids the information loss in TA's multi-step distillation and Meta KD's complex meta-learning process, enabling more efficient training and better parameter adaptation for compact models.

\subsection{Ablation Analysis}
The ablation study highlights significant performance enhancements in Parameter Integration and Teacher-Student Interplay through models KD${stu}$, PI-KD${stu}$, and CrossKD${stu}$, as shown in Table~\ref{tab:combined_performance}. Specifically, the CrossKD${stu}$ model, incorporating both parameter integration and teacher-student dynamics, outperforms others with the highest average scores on the CLUE.
The performance improvements are attributed to the synergistic effects of parameter integration and the dynamic interaction between teacher and student models in the cross-distillation process. While the KD${stu}$ model serves as a baseline, the introduction of parameter integration in PI-KD$_{stu}$ and the combination of both strategies in CrossKD$_{stu}$ progressively enhance model effectiveness.
This underlines the effectiveness of melding parameter integration with teacher-student interplay in cross-distillation, evidencing their combined potential to enhance model performance. Experimental results demonstrating loss convergence superiority are presented in Appendix Sec.~\ref{sec:addexp2}.

\subsection{Effectiveness of Token Pruning and Quantization}
\begin{figure}[!t]
\centering
\includegraphics[width=0.4\textwidth]{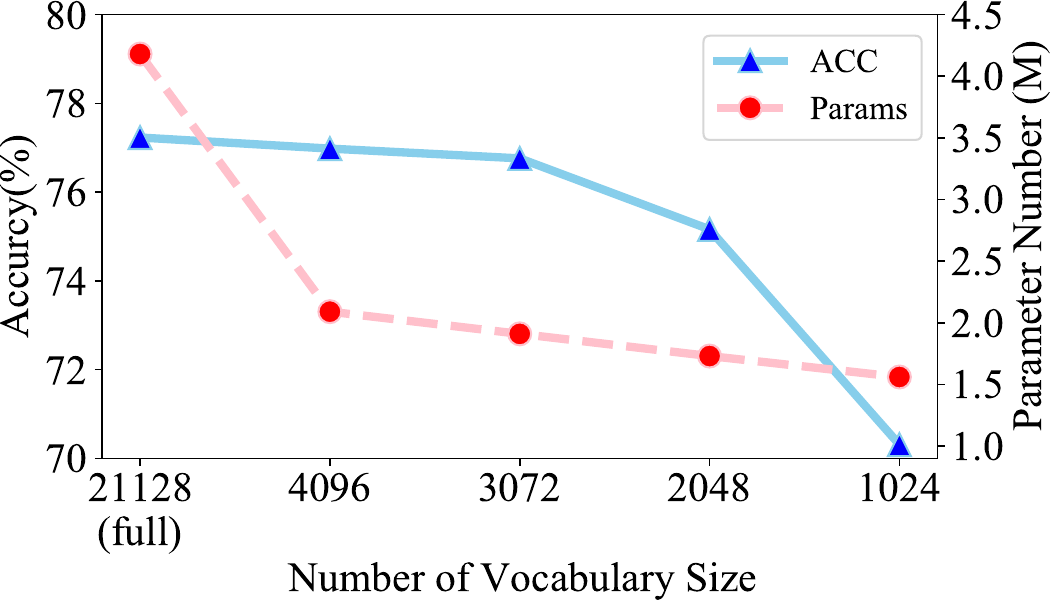}
\vspace{-6pt}
\caption{Hard token pruning evaluation on NER Sector of our collected Alipay Dataset.}
\label{fig:Pruning}
\vspace{-10pt}
\end{figure}
Figure~\ref{fig:Pruning} shows that reducing vocabulary size to 3,072 minimally impacts accuracy, but the further reduction to 2,048 decreases it significantly. CrossKD-TP$_{stu}$ and EI-BERT demonstrate that token pruning reduces parameters effectively. Token pruning efficiently decreases parameters and storage requirements; however, over-pruning leads to a reduction in model performance. 
Achieving an optimal balance in token pruning is essential for sustaining model performance, emphasizing the significance of precise pruning strategies in ultra-model compression. 
Moreover, as indicated by Figure~\ref{fig:flops} and Table~\ref{tab:alipaydata}, the EI-BERT model, following int8 PTQ, shows a slight drop in average accuracy with a significant reduction in storage and operations~(OPs) number compared to its non-quantized counterpart, the CrossKD-TP$_{stu}$ model. Quantization slightly affects accuracy but proves vital for enhancing deployment and efficiency. Thus, employing int8 PTQ is essential for ultra-lite model compression, balancing accuracy and efficiency.

\subsection{Compression Analysis of Our Framework}
We also compared our model with recent baselines in language model compression and distillation. As shown in Table \ref{tab:model_efficiency}, the state-of-the-art methods have not achieved ultra-compression in NLU scenarios. However, the accuracy trade-off of our ultra-compressed EI-BERT model (with TNEWS accuracy of 53.98\% and 1.91MB in storage) is acceptable in practical local plugin-and-play applications.
Furthermore, it is worth noting that the TEB~\cite{liang2023less} method proposes a distillation approach and is currently considered the best task-aware distillation method for language model compression. However, its performance significantly deteriorates when compressed to 17MB. These results highlight the challenge of maintaining efficacy in ultra-compressed models, underscoring the advanced nature of our cross-distillation approach.
According to Figure~\ref{fig:flops}, EI-BERT and CrossKD-TP${stu}$ significantly outperform the Finetuned-BERT${Teacher}$ in terms of compression ratio and inference time speedup. Compared with the original teacher,  CrossKD-TP$_{stu}$ achieves a compression ratio of 53.55$\times$ and an inference speedup of 34.27$\times$, whereas EI-BERT reaches a compression ratio of 213.10$\times$ and an inference speedup of 136.56$\times$. 
The pronounced advantages of EI-BERT underscore its superior performance in compression efficiency and execution speed, proving the importance of implementing model optimization techniques in resource-constrained environments to ensure efficient deployment. To the best of our knowledge, our 1.91MB NLU model is the \textbf{most compact BERT-based language model} deployed at scale in real-world industrial scenarios. 

\begin{table}[!t]
    \centering
    \caption{Comparison of Other Framework}
    \label{tab:model_efficiency}
    \resizebox{0.4\textwidth}{!}{
        \begin{tabular}{@{}lcc@{}}
            \toprule
            Model & \textbf{TNEWS (\%)} & \textbf{Model Storage (MB)} \\
            \midrule
            Bert Base (Teacher) & 57.47 & 407.0 \\
            TinyBert2~\cite{jiao2019tinybert} & 51.08 & 36.7 \\
            TED~\cite{liang2023less} & 51.85 & 17.6 (Albert) \\
            MiniLM~\cite{wang2020minilm} & 54.61 & 68.8 \\
            ERNIE~\cite{liu2023ernie} & 55.30 & 39.1 \\
            MiniRBT~\cite{yao2023minirbt} & 54.47 & 40.9 \\
            EI-BERT(Ours) & 53.98 & 1.91 \\
            \bottomrule
        \end{tabular}
    }
    \vspace{-10pt} 
\end{table}

\section{Online Deployment}
In this section, we present three real mission-critical scenarios at Alipay where the deployment of EI-BERT has demonstrated substantial operational improvements.
\subsection{Real-time Structured Features for EdgeRec}

\begin{figure}[!t]
    \centering
    \begin{subfigure}[b]{0.15\textwidth}
        \includegraphics[width=\textwidth]{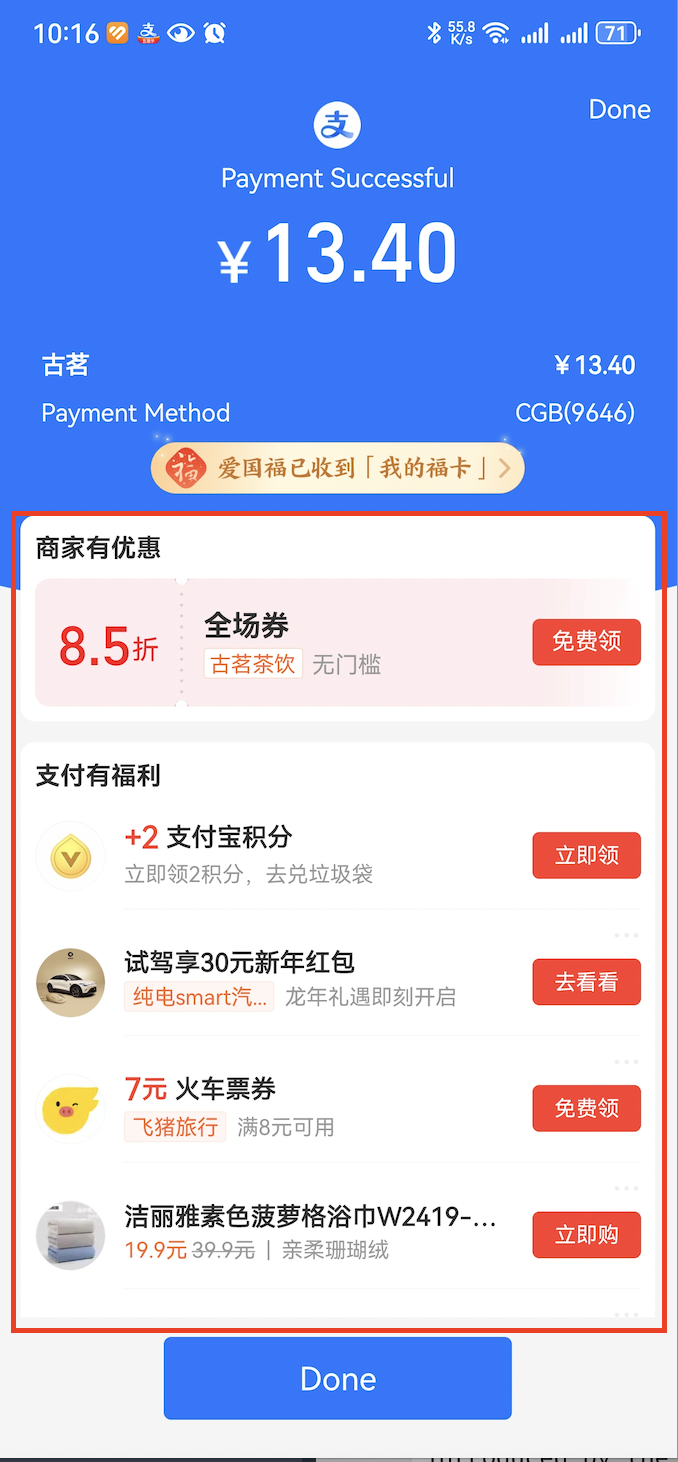}
        \caption{Coupon issuance recommendations.}
        \label{fig:rec2}
    \end{subfigure}
    \begin{subfigure}[b]{0.15\textwidth}
        \includegraphics[width=\textwidth]{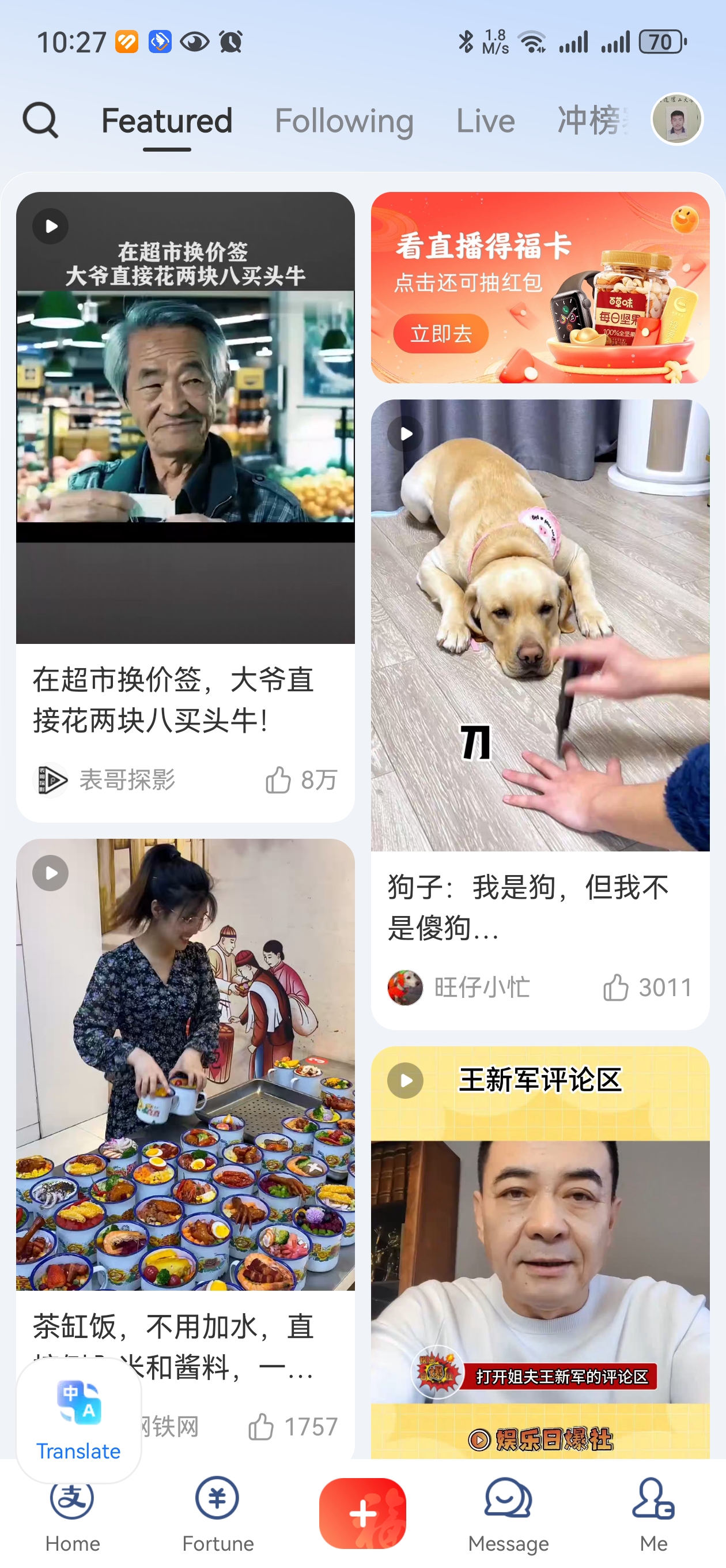}
        \caption{Short video recommendations.}
        \label{fig:recvideos}
    \end{subfigure}
    \begin{subfigure}[b]{0.15\textwidth}
        \includegraphics[width=\textwidth]{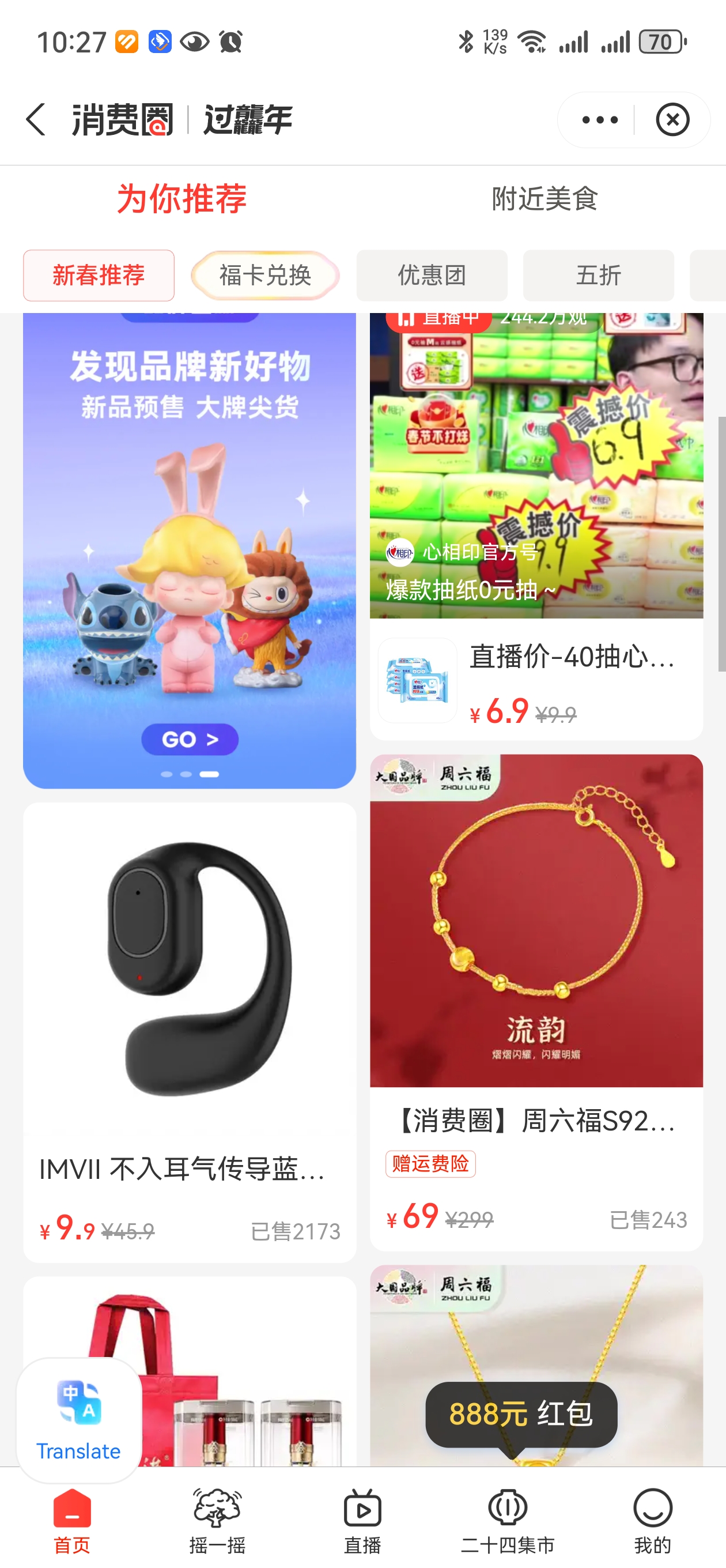}
        \caption{Product recommendations.}
        \label{fig:rec1}
    \end{subfigure}
    \caption{Example Application Scenarios of EI-BERT in Alipay for enhancing user experience through local processing.}
    \label{fig:rec}
\vspace{-10pt}
\end{figure}
As illustrated in Figure~\ref{fig:rec}, our EI-BERT framework has been fully integrated into Alipay's live Edge Recommendation system since January 2024, currently serving the app's recommendation traffic across \textbf{8.4 million daily active devices}. This operational deployment processes 21 million real-time requests per day through on-device execution of multimodal user behavior analysis and item feature extraction, eliminating cloud data transmission while maintaining 95th percentile latency below 80 ms. Post-launch A/B testing over 9.2 million user sessions demonstrated statistically significant improvements in core business metrics: a 4.23\% increase in PV-Click (p=0.0037) and a 3.3\% uplift in PV-CTR (p=0.0082) for coupon recommendations. These production metrics, collected over 30 days of continuous operation, validate the framework's real-world impact beyond experimental simulations, verifying all performance through monitoring systems of Alipay.

\subsection{Edge Intelligent Assistant}
The Intelligent Assistant in Alipay, serving hundreds of millions of users, plays a pivotal role in delivering critical services across finance, commerce, and other domains. Previously, the cloud-based approach for NLU intent recognition exhibited an average latency of ~1s, frequently failing to satisfy Alipay's stringent sub-300ms requirement and lacking offline capability—critical limitations for privacy-sensitive scenarios like healthcare. By deploying EI-BERT's edge-native framework, we achieved a 65\% reduction in latency while enabling fully localized intent recognition. This edge-NLU system now processes sensitive user requests (e.g., medical queries) without cloud dependency, maintaining 98.2\% accuracy parity with its cloud counterpart while ensuring end-to-end data sovereignty. Post-launch metrics over 30 days demonstrate a sustained 214ms average response time across 12.6M daily edge inferences, directly enabling 37 new privacy-constrained use cases in Alipay's production environment.

\subsection{Privacy Edge NLU for Mini-programs}
Alipay's Mini-program ecosystem (spanning healthcare, retail, and property services) now leverages EI-BERT for edge-native natural language understanding, overcoming the privacy limitations of its predecessor’s cloud-dependent architecture. By enabling local execution of NLU tasks including contextual query resolution, service categorization, and sensitive data detection, the framework eliminates cloud data transmission while maintaining 99\% parity with cloud model accuracy. Post-deployment metrics from 30 days of live operation across 18 million daily active instances demonstrate an 85\% reduction in network payload and a 40\% decrease in cloud compute costs through edge offloading. Critically, privacy-sensitive verticals like medical Mini-programs showed a significant 12.3\% improvement in 30-day user retention rates (p<0.01).

\section{Related Works}
\noindent\textbf{Knowledge Distillation.}
Recent advances in knowledge distillation~\cite{hinton2015distilling, chen2021cross} have significantly enhanced the efficiency of language models~\cite{wang2022deep}, with models like TinyBERT~\cite{jiao2019tinybert} and DistilBERT~\cite{sanh2019distilbert} achieving substantial size reductions without compromising performance. Addressing capacity disparity approaches such as Teacher Assistant (TA)~\cite{mirzadeh2020improved}, MiniDisc~\cite{zhang2022minidisc}, and MiniMoE~\cite{zhang2023lifting} utilize multi-step distillation and expert mechanisms, yet they often introduce increased complexity and higher resource requirements. In the realm of teacher adaptivity, methods like Meta KD~\cite{zhou2021bert} and Sparse Teacher~\cite{yang2022sparse} optimize teacher models through meta-learning and parameter reduction, respectively, but may add training complexity or struggle with complex tasks. Despite these advancements, balancing complexity and efficiency remains a challenge, especially for ultra-compressed models (under 5MB). Our proposed cross-model distillation method addresses these issues by streamlining knowledge transfer and enabling dynamic adaptation, thereby reducing implementation complexity
and enhancing efficiency.

\noindent\textbf{On-Device Compression Framework.}
In the domain of on-device model deployment, particularly in the context of edge computing, there have been notable advancements. The EdgeRec system serves as a prime example, demonstrating this progression with its mobile device recommender system that intelligently adapts to user preferences, signifying a substantial advancement in on-device personalization~\cite{gong2020edgerec}. Concurrently, there has been considerable progress in model compression techniques, especially for transformer models. A prominent instance is the deeply compressing pre-trained transformer models, which significantly reduces the model size while preserving accuracy, a vital consideration for deploying complex models in resource-limited environments~\cite{wang2022deep}. Similarly, the approach to efficient model deployment on edge devices highlights the ongoing efforts to enhance computational efficiency within resource constraints~\cite{kharazmi2023distill}. Due to limited capabilities, existing industrial frameworks have not yet achieved plug-and-play ultra-compressed models that maintain competitive performance.

\section{Conclusion}
In this paper, we presented EI-BERT, an edge ultra-lite BERT framework with a novel cross-distillation method, addressing the fundamental challenges of deploying NLP models in resource-restricted mobile environments. Our framework effectively tackles the dual requirements of extremely small model sizes and strong knowledge transfer capabilities through a comprehensive compression pipeline featuring cross-distillation as its cornerstone. The effectiveness of our approach is demonstrated through achieving an ultra-compact NLU model of just \textbf{1.91MB} - the smallest known BERT-based language model to date. This remarkable compression enables secure local processing and real-time responses on mobile devices while maintaining robust performance across diverse NLP tasks. Our framework has been successfully implemented across numerous real-world language understanding scenarios within the Alipay ecosystem.
A detailed discussion of limitations and future work can be found in Appendix Sec.~\ref{sec:Limitations}.

\balance
\bibliographystyle{ACM-Reference-Format}
{\bibliography{main}}
\begin{appendices}

\section{Details of Alipay Datasets}
\label{appendix:Alipay}
\begin{figure}[!h]
  \centering
  \includegraphics[width=0.45\textwidth]{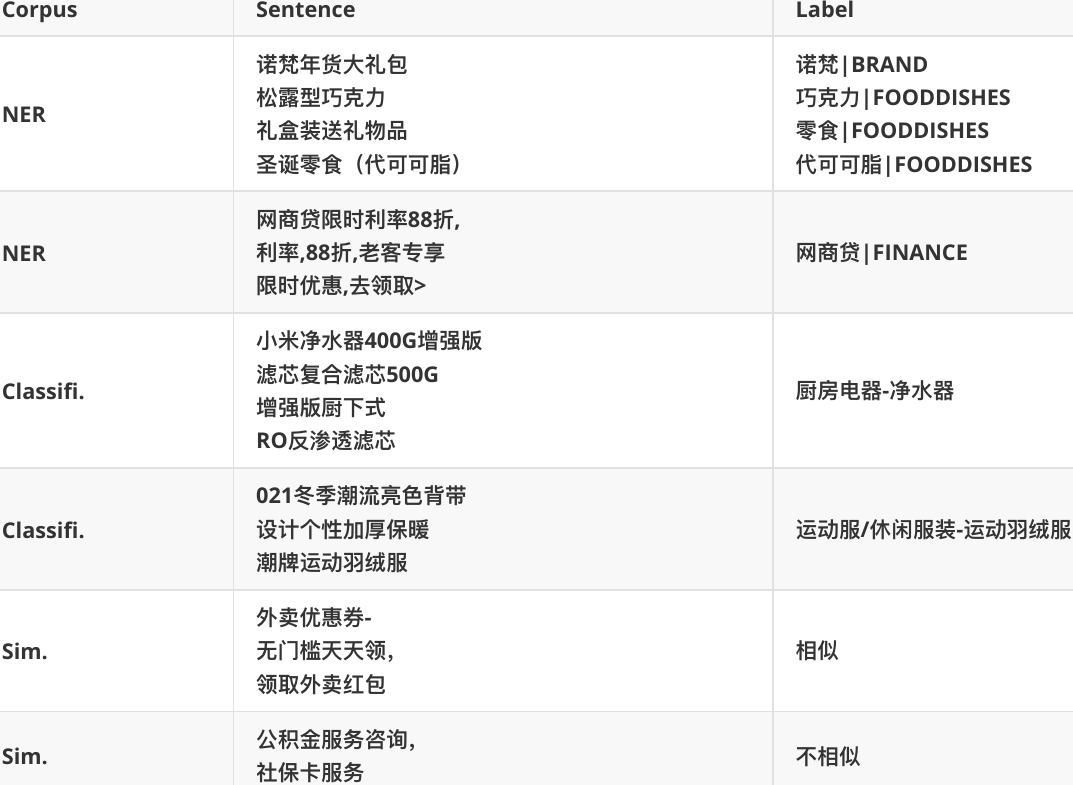}
  \caption{Some specific examples of Alipay Datasets.}
  \label{fig:specific}
\end{figure}

 Alipay Datasets encompass a diverse range of tasks. Named Entity Recognition, which is evaluated using the F1-score, identifies entities such as brands, products, and financial entities across 29 major entity types, with data sourced from Alipay's marketing and financial datasets. Classification tasks assess 31 primary and 108 secondary categories, using accuracy as the primary metric for evaluation. Additionally, the dataset includes Similarity tasks, which are crucial for determining the likeness between different data points.
Table~\ref{tab:alipaysta} displays various fundamental statistics, and Figure~\ref{fig:specific} illustrates some basic examples.
\begin{table}[!h]
\centering
\caption{Dataset Summary for Model Validation}
\label{tab:alipaysta}
\resizebox{0.5\textwidth}{!}{%
\begin{tabular}{@{}llllll@{}}
\toprule
Corpus & Train & Dev & Test & Metric & Source \\
\midrule
\textbf{NER Dataset} & 47.9k & 2.8k & 6.8k & F1 & Alipay Marketing, Financial Data \\
\textbf{Classification Dataset} & 14.6k & 858 & 2.1k & Accuracy & Alipay Marketing, E-commerce Data \\
Similarity & 7.5k & 444 & 1.1k & Accuracy & Mini-program, E-commerce, Marketing \\
\bottomrule
\end{tabular}%
}
\end{table}

\section{Baseline Details}
\label{sec:base}
We choose our carefully finetuned BERT$_{base}$~\cite{devlin2018bert} as teacher model. For our NLU student model, we chose a configuration that optimizes both efficiency and effectiveness. Our model architectures align with ALBERT$_2$~\cite{lan2019albert} and include:
a.~\textit{ALBERT$_{2}$ (Student)} enhances the base model through finetuning with a specialized corpus, improving adaptability and performance. b.~\textit{KD$_{stu}$} employs traditional knowledge distillation techniques to further advance the model's capabilities.
c.~\textit{PI-KD$_{stu}$} incorporates parameter integration for model knowledge distillation. d.~\textit{CrossKD$_{stu}$} adopts full process of cross-distillation. Building further, e.\textit{CrossKD-TP$_{stu}$} includes hard token pruning for reduced model storage. Lastly, f.~{EI-BERT}, which stands for \textbf{E}xtremely L\textbf{I}ght BERT, represents the most advanced model, combining previous features with 8-bit quantization. Additionally, we've also selected two compact models for comparison: \textbf{TinyBERT}~\cite{jiao2019tinybert}, which is adapted for processing the Chinese language with notable efficiency in resource utilization, and \textbf{ALBERT}~\cite{lan2019albert}, chosen for its parameter efficiency attained through cross-layer parameter sharing, with ALBERT$_4$ and ALBERT$_2$ serving as specific baselines.

\begin{table*}[!h]
    \centering 
    \caption{Comparison of TNEWS Task-specific Loss Values Across Epochs in TNEWS}
    \label{tab:comparison_4}
    \resizebox{\textwidth}{!}{
        \begin{tabular}{@{}cccccccccc@{}} 
            \toprule 
            Model & epoch0 & epoch1 & epoch2 & epoch3 & epoch4 & epoch5 & epoch6 & epoch7 \\
            \midrule
            Student (Ordinary Knowledge Distillation) & 2.3450 & 1.7818 & 1.4137 & 1.3185 & 1.2355 & 1.1737 & 1.1446 & 1.1217 \\
            Student (cross-distillation) & 1.7380 & 1.4860 & 1.2725 & 1.1075 & 1.1737 & 1.0140 & 0.7685 & 0.6324 \\
            \bottomrule
        \end{tabular}
    }
\end{table*}

We comprehensively evaluate our approach against several state-of-the-art distillation methods that represent different paradigms in model compression and knowledge transfer. Task-aware layEr-wise Distillation (TED)\cite{liang2023less} represents a sophisticated approach that utilizes task-aware filters to achieve precise alignment between the hidden representations of student and teacher models. This method's distinctive feature lies in its ability to selectively transfer task-relevant knowledge, ensuring that the student model acquires the most pertinent information for specific tasks. We also implement the Teacher Assistant (TA) framework\cite{mirzadeh2020improved} with two different configurations: one using TinyBERT${4}$ as the teaching assistant and another employing ALBERT${4}$. This multi-step distillation approach aims to bridge the significant architectural gap between teacher and student models through intermediate models, facilitating more effective knowledge transfer. The TA method's use of intermediate models helps manage the complexity gap between the original teacher and final student models, potentially leading to better knowledge distillation outcomes. Additionally, we compare against Meta KD~\cite{zhou2021bert}, an advanced approach that leverages meta-learning to dynamically adjust the teacher's guidance during the knowledge distillation process. While Meta KD offers sophisticated optimization of the teacher model to enhance student learning, it introduces additional complexity to the training pipeline through its meta-learning mechanisms. This method focuses on finding optimal teaching strategies by adapting the knowledge transfer process based on the student's learning progress and needs.

\section{Loss Convergence Comparison}
\label{sec:addexp2}
As shown in Table~\ref{tab:comparison_4}, the cross-distillation enhanced model consistently reduces loss more effectively than the ordinary knowledge distillation-based procedure. Early in training (epochs 0 to 3), the loss decreases faster, and by the final epochs (5 to 7), the cross-distillation model achieves a much lower final loss (0.6324 compared to 1.1217).
This is because cross-distillation addresses the capacity disparity between teacher and student models and adapts knowledge transfer to the unique learning requirements and environmental contexts of ultra-compact student models. Ordinary knowledge distillation, relying on a single static teacher model, struggles with these challenges, leading to less effective learning.
These results demonstrate the necessity of cross-distillation for ultra-compact model compression.

\section{Limitations and Future Works}
\label{sec:Limitations}
We acknowledge several limitations in our cross-distillation and compression framework, particularly when applied to personalized applications and ultra-compact models~(sub 5MB) for natural language generation (NLG) tasks.
Our current framework involves distilling a single model for a specific task and then deploying it to all users without considering their unique characteristics. The lack of user-specific information during the distillation process is another limitation of our approach.
Moreover, our framework still requires retaining the teacher model’s task-specific head during deployment to maintain performance quality. This architectural dependency imposes non-trivial constraints on deployment flexibility and prevents complete decoupling from the original teacher model structure.

Looking ahead, we see opportunities to extend our framework for edge AI challenges by pursuing two interconnected strategies. First, we aim to bridge the gap between large language models (LLMs) and edge deployment on ~\textbf{extremely resource-constrained conditions} through ultra-compact knowledge distillation, focusing on preserving core LLM capabilities—such as multi-step reasoning, in-context learning, and structured knowledge representation—under extreme resource constraints. This involves developing dynamic encoding methods to compress reasoning patterns and knowledge bases while maintaining functional fidelity. Second, we are advancing parameter-efficient adaptation by designing modular interfaces that enable rapid domain transfer through selective updates. By freezing distilled core models and only fine-tuning lightweight task-specific components, the system retains computational efficiency while adapting to new scenarios. This dual approach addresses both the fundamental constraints of edge hardware and the practical need for adaptable AI systems.

Furthermore, we aim to enhance our framework's ability to handle more complex scenarios through advanced knowledge transfer techniques and adaptive compression strategies that better balance model size, inference speed, and task performance. Through these future directions, we envision pushing the boundaries of what's possible with edge AI, working toward a future where sophisticated AI capabilities are readily available on edge devices while maintaining privacy, efficiency, and performance. This vision includes developing more sophisticated approaches to incorporate user-specific information and creating more adaptive, personalized models while maintaining the benefits of edge deployment.

\end{appendices}
\end{document}